\documentclass[twoside,11pt]{article}

\usepackage{jmlr2e}

\usepackage{amsmath}
\usepackage{mathtools}

\RequirePackage{adjustbox}[2019/01/04]
\usepackage[utf8]{inputenc} 
\usepackage[T1]{fontenc}    
\usepackage{hyperref}       
\usepackage{url}            
\usepackage{booktabs}       
\usepackage{amsfonts}       
\usepackage{nicefrac}       
\usepackage{microtype}      
\usepackage{xcolor}         
\usepackage{graphicx}
\usepackage{bm}
\usepackage{enumitem}

\usepackage{graphicx} 

\newcommand{\E}{\mathbb{E}}
\newcommand{\R}{\mathbb{R}}
\newcommand{\HV}{\text{HV}}

\newcommand{\EHVI}{\text{EHVI}}

\newcommand{\bbX}{\ensuremath{\mathbb{X}}}

\newcommand{\calN}{\ensuremath{\mathcal{N}}}
\newcommand{\calP}{\ensuremath{\mathcal{P}}}
\newcommand{\calS}{\ensuremath{\mathcal{S}}}
\newcommand{\calY}{\ensuremath{\mathcal{Y}}}

\newcommand{\strength}{\texttt{strength}}
\newcommand{\GWP}{\texttt{GWP}}

\newcommand{\x}{{\mathbf x}}
\newcommand{\X}{{\mathbf X}}
\newcommand{\y}{{\mathbf y}}
\newcommand{\f}{{\mathbf f}}

\newcommand{\refpoint}{{\mathbf r}}

\newcommand{\coo}{\ensuremath{\mathrm{CO_2}}}

\title{
Sustainable Concrete via 
Bayesian Optimization
}
\date{September 2023}

\author{%
  Sebastian Ament\textsuperscript{1}
  \hfill \texttt{ament@meta.com} \\
  Andrew Witte\textsuperscript{2}
  \hfill \texttt{dreww2@illinois.edu} \\
  Nishant Garg\textsuperscript{2}
  \hfill \texttt{nishantg@illinois.edu} \\
  Julius Kusuma\textsuperscript{1}
  \hfill \texttt{jkusuma@meta.com} \\
    {
    \normalfont
    \textsuperscript{1}Meta 
    and
    \textsuperscript{2}University of Illinois at Urbana-Champaign.
    }
}

\begin{document}

\maketitle

\begin{abstract}
Eight percent of global carbon dioxide emissions can be attributed to the production of cement, the main component of concrete, which is also the dominant source of \coo{} emissions in the construction of data centers.
The discovery of lower-carbon concrete formulae is therefore of high significance for sustainability.
However, experimenting with new concrete formulae is time consuming and labor intensive,
as one usually has to wait to record the concrete's 28-day compressive strength,
a quantity whose measurement can by its definition not be accelerated.
This provides an opportunity for 
experimental design methodology like Bayesian Optimization (BO) to accelerate the search for strong and sustainable concrete formulae.
Herein, we 1) propose modeling steps that make concrete strength amenable to be predicted accurately by a Gaussian process model with relatively few measurements,
2) formulate the search for sustainable concrete as a multi-objective optimization problem,
and 3) leverage the proposed model to carry out multi-objective BO
with real-world strength measurements of the algorithmically proposed mixes.
Our experimental results show improved trade-offs between the mixtures' global warming potential (GWP) and their associated compressive strengths, compared to mixes based on current industry practices.
Our methods are open-sourced at \href{https://github.com/facebookresearch/SustainableConcrete}{\texttt{github.com/facebookresearch/SustainableConcrete}}.
\end{abstract}

\section{Introduction}

Eight percent of global carbon dioxide emissions can be attributed to the production of cement~\citep{lehne2018making}, the main reactive component of concrete, 
contributing significantly to anthropogenic climate change~\citep{solomon2009irreversible}.
By comparison, the global annual emission from commercial aviation was estimated at $2.4\%$ in 2019~\citep{graver2019emissions}.
Concrete is also the leading source of \coo{} emissions in data center construction,
accounting for 20-30\% of the associated emissions,
and making the reduction of the carbon footprint of concrete necessary 
to de-carbonize the operations of modern technology companies.
Further, concrete mixtures that simultaneously exhibit a small carbon footprint and safe strength levels could become a critical piece in achieving societal de-carbonization goals and the mitigation of climate change.


However, conventional concrete is mainly optimized for cost, availability, and compressive strength at the 28-day mark. To meet construction and sustainability goals, concrete needs to be optimized for additional, often opposing objectives: curing speed and low environmental impact,
where the latter is commonly expressed as the global warming potential (GWP), typically in kilo-gram of \coo{} per cubic meter.
The optimization of these opposing objectives is the primary goal of this work,
and is part of a program to develop low-carbon concrete for data center construction which includes model development, lab testing, pilot projects, and at-scale application at Meta's data centers~\citep{ge2022accelerated, greenbuild2022concrete}.

Herein, we give an overview of our methodology and validated experimental results 
that enable reliable strength predictions and the optimization of the trade-offs that are inherent to the design of low-carbon concrete.
In particular, we 
1) propose a probabilistic model that maps concrete formulae to compressive strength curves, 
2) formulate the search for sustainable concrete as a multi-objective optimization problem, and
3) employ Bayesian optimization (BO) in conjunction with the proposed model to accelerate the optimization of the strength-GWP trade-offs using {\it real-world} compressive strength measurements.

\section{Background}

\subsection{Gaussian Processes}
\label{subsec:background:GPs}

Gaussian Processes (GP) constitute a general class of probabilistic models that permit exact posterior 
inference using linear algebraic computations alone~\citep{Rasmussen2004},
which includes the quantification of the uncertainty of the model predictions.
For this reason, most BO approaches use GPs as a model for the objective function $f$ that is to be optimized and presumed to be too expensive to evaluate frequently.
GPs can be defined as a class of distributions over functions whose finite-dimensional marginals are multi-variate Normal distributions.
Formally, a real-valued $f$ is distributed according to a GP 
if for any set of points $\bm X = [\bm x_1, \hdots, \bm x_n]$ in the domain $\bbX$,
\begin{equation}
\label{eq:gp_definition}
    f(\X) = [f(\x_1), \hdots, f(\x_n)] \sim \calN(\mu(\X), k(\X, \X)), 
\end{equation}
where 
$\mu: \bbX \rightarrow \R$ is called the mean function and
$k: \bbX \times \bbX \rightarrow \R$ the covariance function, or simply ``kernel'',
and we overloaded the notation of $f, \mu$ and $k$ applied to the {\it set} of inputs $\X$ with a ``broadcasting'' over the elements of $\X$:
$\mu(\X) = [\mu(\x_1), \hdots, \mu(\x_n)]$ and $k(\X, \X')_{ij} = k(\x_i, \x'_j)$ is a matrix which is positive semi-definite if $\X = \X'$.

\subsection{Multi-Objective Bayesian Optimization}
\label{sec:mobo}

Multi-objective optimization (MOO) problems 
generally exhibit trade-offs between $m > 1$ objectives $\f = [f_1, \cdots, f_m]$ that make it impossible to find a single input that jointly maximizes the objectives.
Instead, one is usually interested in finding a set of optimal {\it trade-offs}, also called the Pareto frontier (PF) between multiple competing objectives,
usually under the constraint that each objective is 
above a minimum acceptable value $f_i(\x) > r_i$.
Collectively, the set of lower bounds $\refpoint = [r_1, \cdots, r_m]$ is referred to as the {\it reference point}.


The {\it hypervolume} of a discrete Pareto frontier $\calP = \{\y_i\}_i$ bounded by a reference point $\refpoint$
is a common measure of the quality, formally 
 $\HV(\calP, \refpoint):= \lambda\bigl(\bigcup_{\y_i \in \calP} [\refpoint, \y_i]\bigr)$, where $[\refpoint, \y_i]$ denotes the hyper-rectangle bounded by vertices $\refpoint$ and $\y_i$, and $\lambda$ is the Lebesgue measure. 
Thus, a natural acquisition function for MOO problems is the expected hypervolume improvement
\begin{equation}
    \begin{aligned}
    \label{eq:intro:qEHVI}
    \EHVI(\x) 
    = \E \left[ [\HV(\calP \cup \f(\x), \refpoint) - \HV(\calP, \refpoint)]_+ \right],
    \end{aligned}
\end{equation}
from obtaining a set $\calY \sim \f(\X) := [\f(\x_1), \cdots, \f(\x_q)]$ of $q$ new observations. 
If $q=1$ and the objectives are modeled with independent GPs, {\EHVI} can be expressed in closed form~\citep{yang2019mobgoehvi},
otherwise Monte Carlo approximations are necessary \citep{daulton2020ehvi}. 

Notably, \citet{derousseau2021mooforlowcarbonconcrete} proposed a simulation-optimization framework for low-carbon concrete
which jointly optimizes deterministic models of 28-day strength, cost, and embodied carbon
using using an evolutionary algorithm (EA).
The approach is promising but unlikely to be competitive for real-world experimentation without modifications, as EAs tend to be significantly less sample efficient than BO.


\section{A Probabilistic Model of Compressive Strength}
\label{sec:probabilistic_model}

In prior work,
\citet{carino2001maturity_method} proposed analytical forms for the evolution of concrete's compressive strength as a function of time, \citet{paal2021concrete_svm} proposed a non-temporal, non-probabilistic kernel regressor to predict concrete's lateral strength,
and \citet{ge2022accelerated} used conditional variational autoencoders to predict the compressive strength at discrete $t$-day intervals. 
Herein, we propose a probabilistic model $\strength(\x, t)$
that jointly models dependencies on composition $\x$ and time $t$.
The proposed model is accurate even in the low data regime due to
data transformations, augmentations, and a customized kernel function.

\paragraph{Zero-day zero-strength conditioning}

A simple but key characteristic of concrete strength is that it is zero at the time of pouring.
As no actual measurement is made upon pouring,
strength data sets generally do not contain records of the zero-day behavior.
Direct applications of generic machine learning models to these data sets can consequently
predict non-physical, non-sensical values close to time zero.

We propose a simple data augmentation that generalizes to any model type to fix this:
For any concrete mixture in a training data set, we add an artificial observation at time zero with corresponding strength zero. 
We add additional artificial observations at time zero for a randomly chosen subset of compositions $\x$ to encourage the model to conform to the behavior for compositions that are dissimilar to any observed ones.

\paragraph{Non-stationarity in time}
The evolution of compressive strength is non-stationary in time,
as the strength increases quickly and markedly early in the curing process, but converges monotonically to a terminal strength value.
Therefore, we transform the time dimension to be logarithmic before passing the inputs to the model: $t \to \log t$ .
This transformation enables the application of stationary covariance kernels on the log-time dimension, while leading to good empirical predictive performance.

\paragraph{Kernel design}
Aside from data augmentation and transformation,
a GP's kernel chiefly determines the behavior of the model and permits the incorporation of problem-specific, physical structure~\citep{ament2021sara}.
Notably, the $\strength(\x, \cdot)$ curves generally share a similar shape for any composition $\x$.
For this reason, we introduce an additive composition-independent time-dependent component $k_{\text{time}}$ to a generic kernel over all variables $k_{\text{joint}}$:
\begin{equation}
    k((\x, t), (\x', t')) = \alpha k_{\text{time}}(\log t, \log t') + \beta k_{\text{joint}}((\x, \log t), (\x', \log t')),
\end{equation}
where $\alpha, \beta > 0$ are variance parameters which are inferred with the other kernel hyper-parameters
using marginal likelihood optimization~\citep{Rasmussen2004}.
For our experiments, we chose $k_{\text{time}}$ to be an exponentiated quadratic kernel 
and $k_{\text{joint}}$ to be a Mat\'ern-5/2 kernel with automatic relevance determination (ARD).
This structure allows the model to infer the general smooth shape of strength curves independent of the particular composition $\x$, leading the model's predictions to be physically sensible and accurate for long time horizons.

\paragraph{Model Evaluation}

\begin{figure}[t!]
    \includegraphics[width=\textwidth]{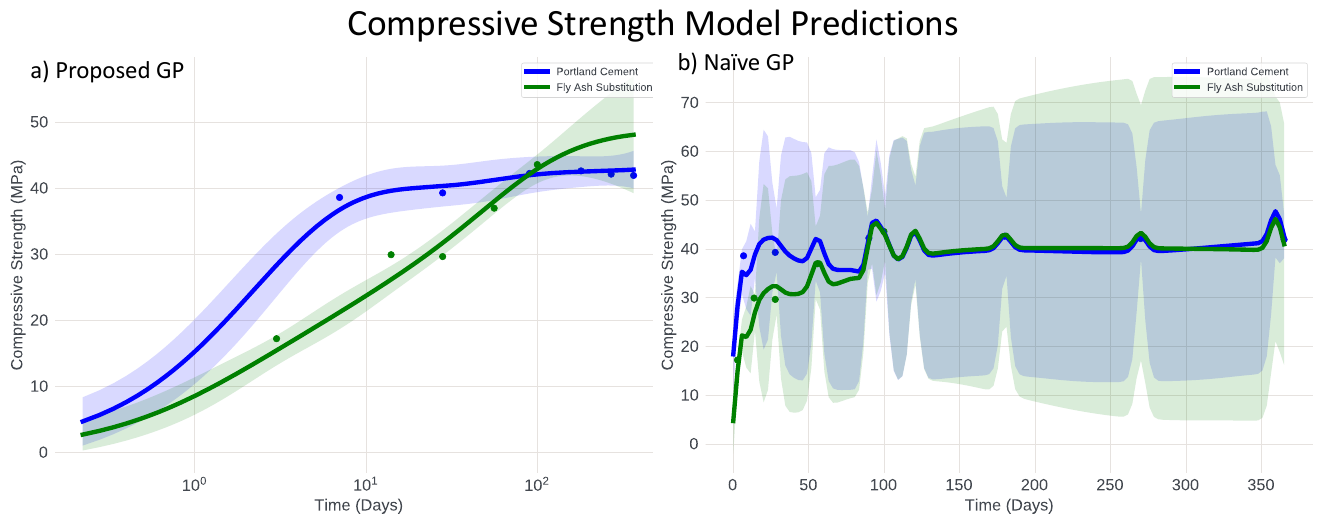}\vspace{-1.5ex}
    \caption{Strength curve predictions 
    by our proposed model (left) and a standard GP (right)
    for two different compositions $\x$: 1) 100\% cement (blue) and 2) 20\% fly-ash (green).
    }
    \label{fig:experiments:strength_curves}
\end{figure}

Figure~\ref{fig:experiments:strength_curves}
shows the predicted strength curves $\strength(\x, \cdot)$
for two fixed compositions $\x$, but ranging over time $t$,
of our model (left) and a na\"ive GP (right).
Both models were trained on the UCI concrete strength dataset~\citep{misc_concrete_compressive_strength_165},
which we used to develop our model before starting 
our own experiments.
While a na\"ive application of a GP to the data (dots) is unsuccessful,
our model is accurate and well calibrated. 

\section{Sustainable Concrete as a Multi-Objective Optimization Problem}
\label{sec:moo_formulation}

Increasing sustainability is the motivating factor of this work, though it is simultaneously critical to maintain concrete's compressive strength above application-specific thresholds.

Specifically, our primary objective is making concrete {\it sustainable},
a multi-faceted notion that includes, but is not limited to the carbon impact of production.
Here, we focus on the global warming potential (GWP) \cite{} to quantify concrete mixtures' sustainability, but the methods are general and can be extended to metrics that quantify complementary aspects of sustainability.
Importantly, any decrease in GWP would be rendered meaningless by the inability of the associated concrete formula to be used for construction with usually tight deadlines. We therefore add compressive strength at short 1-day and long 28-day curing durations to our list of objectives.

The strength objectives are a-priori unknown and are estimated using the model proposed in Section~\ref{sec:probabilistic_model} by evaluating $\strength(\x, 1)$ and $\strength(\x, 28)$.
We model GWP as a deterministic linear function $\GWP(\x) = \bm \alpha^\top \x$ where each $\alpha_i$ quantifies the GWP for each unit of the $i$th mixture ingredient.
A more precise quantification of GWP would also dependent on location and transportation~\citep{kim2022openconcrete},
but the linear model is a reasonable approximation for our purposes.
Further, given an accurate strength model, we can infer Pareto-optimal mixes
for a post-hoc change in the GWP model, similar to the post-hoc changes in the constraints explored in Sec.~\ref{sec:experiments:inferred_pareto}.
Formally, the associated MOO problem is
\begin{equation}
\label{eq:sustainable_concrete_moo}
    \max_{\x \in \bbX} \ (\strength(\x, 1), \ \strength(\x, 28), \ -\GWP(\x)).
\end{equation}
We then employ BO with \citet{ament2023logei}'s 
qLogNEHVI acquisition function --
the LogEI variant of \citet{daulton2020ehvi}'s qNEHVI --
to design batches of compositions $\X$,
optimizing for the PF of the MOO problem of Eq.~\eqref{eq:sustainable_concrete_moo}
using real-world compressive strength experiments. 

\section{Experiments}
\label{sec:experiments}

\subsection{Real-World Experimental Setup}
All experimental testing of mortar specimens was performed in accordance with ASTM C109, as outlined below. Two-inch mortar cubes were mixed and cured at 22 °C, first mixing fine aggregate with half the water and then adding all cementitious material with the remaining water. Superplasticizer was added during the second stage of mixing, as needed. After mixing and tamping, plastic lining was applied to each mold to prevent significant moisture loss. After 24 hours of curing inside the molds, mortar cube specimens were removed and submerged in a lime-saturated water bath at room temperature (22 °C). Three specimens each were prepared for curing ages of one, three, five, and twenty-eight (1, 3, 5, and 28) days. All specimens were subjected to a compressive load at a constant loading rate of 400 lb/s using a Forney compressive testing machine.

\subsection{Empirical Optimization Results}

\begin{figure}[t!]
    \includegraphics[width=\textwidth]{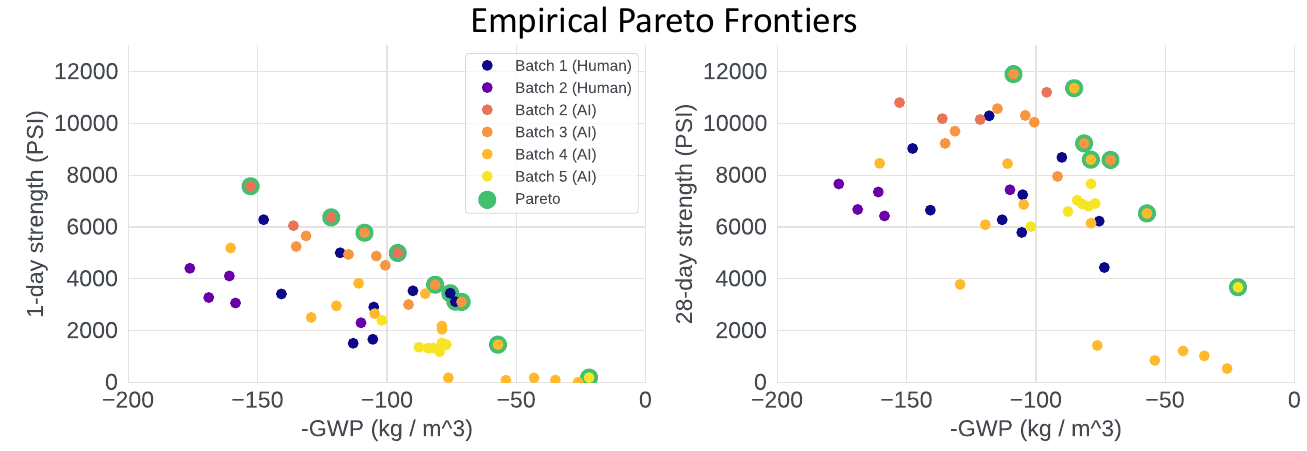}\vspace{-1.5ex}
    \caption{Empirical PFs for 1-day (left) and 28-day (right) compressive strength and GWP. The Pareto-optimal measurements are circled green. 
    The darker colors correspond to human-generated mixes, 
    while colors on the spectrum from dark orange to yellow 
    correspond to AI-proposed mixes.
    }
    \label{fig:experiments:experimental_pareto}
\end{figure}

Figure~\ref{fig:experiments:experimental_pareto} shows the experimentally achieved trade-offs between 1-day (left) and 28-day (right) compressive strength, and GWP.
The proposed mixtures quickly improve on both a random initial set (Batch 1, human)
and a set of mixtures inspired by industry practices (Batch 2, human).
The AI-proposed batches push the empirical Pareto frontier outward, providing an increasingly fine set of trade-offs between sustainability and strength.
Fortunately, 
the highest-GWP Pareto-optimal composition shown in Figure~\ref{fig:experiments:experimental_pareto}
dominates the GWP-strength trade-off of 
a pure-cement mix (not shown in Fig.~\ref{fig:experiments:experimental_pareto}), implying that a GWP-reduction can -- to a non-negligible extent -- be achieved without sacrificing strength, though greater GWP reductions do require such trade-offs.

\subsection{Inferred Pareto Frontier}
\label{sec:experiments:inferred_pareto}
\begin{figure}[t!]
    \includegraphics[width=\textwidth]{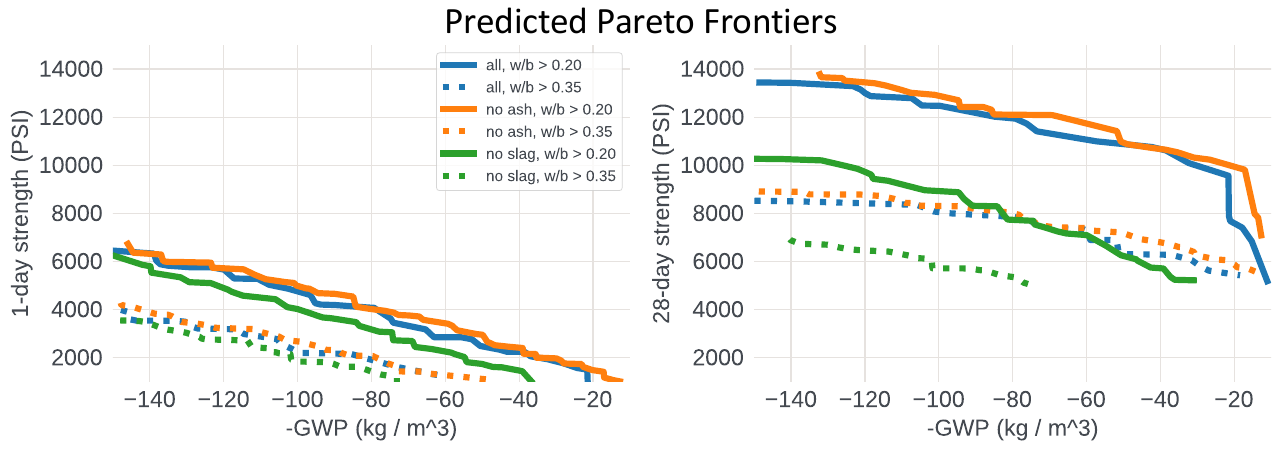}\vspace{-1.5ex}
    \caption{Predicted PFs for 1-day (left) and 28-day (right) compressive strength and GWP.}
    \label{fig:experiments:predicted_pareto}
\end{figure}

In addition to using the proposed concrete strength model for BO,
we can use it to compute inferred PFs conditioned on application-specific constraints.
Querying the model in this way is useful both to gain scientific insight and 
because ingredients like fly ash or slag -- waste-products of coal and steel plants, respectively -- might only be effectively sourced in specific regions. 
By computing the inferred PF based on location-specific constraints,
we can generate composition recommendations that are {\it customized to specific construction projects}.

Figure~\ref{fig:experiments:predicted_pareto} shows a numerical approximation of the inferred PF
conditioned on different water-to-binder (w/b) ratios, 
which affect the workability of the concrete, with higher w/b usually being more workable,
as well as constraining the mixtures not to include fly ash (orange) or slag (green).
The inferred PFs yield several insights: 
1) the water-to-binder ratio has a significant effect on compressive strength, a phenomenon that has been remarked by in the literature on concrete \citep{najaraj1996abramslaw, rao2001abramslawmortar},
2) removing ash from the composition space has a negligible effect on the achievable trade-offs up to 28 days, and 
3) removing slag has a significant negative effect.
We stress that these results are based on predictions, 
but qualitatively match expert consensus on the variables' effects.

\section{Conclusion}

We introduced a probabilistic model for the temporal evolution of concrete's compressive strength as a function of its composition,
formalized the problem of finding strong yet sustainable concrete formulae as a multi-objective optimization problem,
and leveraged BO to propose new concrete mixtures for real-world testing.
We seek to accelerate the development of sustainable concrete 
by open-sourcing our methods at
\href{https://github.com/facebookresearch/SustainableConcrete}{\texttt{github.com/facebookresearch/} \texttt{SustainableConcrete}}.
The work has the potential to decrease the carbon footprint of data center construction and the construction industry at large, with possibly global impact.

\bibliography{concrete}

\newpage 

\appendix 

\section{Background}

\paragraph{Additional Information on Gaussian Processes}
Given a set of input and output pairs $(\X, \y)$ and a Gaussian likelihood with zero mean and variance~$\sigma^2$, the posterior distribution of a GP is also a GP defined by the posterior mean $\mu_p$ and kernel $\Sigma_p$:
\begin{equation}
\label{eq:gp_posterior}
\begin{aligned}
    \mu_p(\X^*) &= \mu(\X^*) + \Sigma(\X^*, \X)(\Sigma(\X, \X) + \sigma^2 \mathbf I)^{-1} (\y - \mu(\X)), \\
    \Sigma_p(\X^*, \X'^*) &= \Sigma(\X^*, \X'^*) - \Sigma(\X^*, \X) (\Sigma(\X, \X) + \sigma^2 \mathbf I)^{-1} \Sigma(\X, \X'^*). \\
\end{aligned}
\end{equation}
These equations show that, without a-priori knowledge about the specific values of the target function, the kernel function $\Sigma$ is the primary factor in determining the behavior of the model. 
In the literature, $\Sigma$ is usually also a function of hyper-parameters like length-scales which control how quickly the function is expected to vary with respect to specific input dimensions. In this work, we use well-established marginal likelihood optimization to optimize our model's hyper-parameters.

In the single-outcome ($M=1$) setting, $\f(\x) \sim \calN(\bm \mu(\x), \bm \Sigma(\x))$ with $\bm \mu: \bbX^q \rightarrow \R^q$ and $\bm \Sigma: \bbX^q \rightarrow \calS^q_+$.

In the sequential ($q=1$) case, this further reduces to a univariate Normal distribution: $f(\x) \sim \calN(\mu(\x), \sigma^2(\x))$ with $\mu: \bbX \rightarrow \R$ and $\sigma: \bbX \rightarrow \R_+$. 

\section{Probabilistic Compressive Strength Model}  

\begin{figure}[t!]
\centering
    \includegraphics[width=0.8\textwidth]{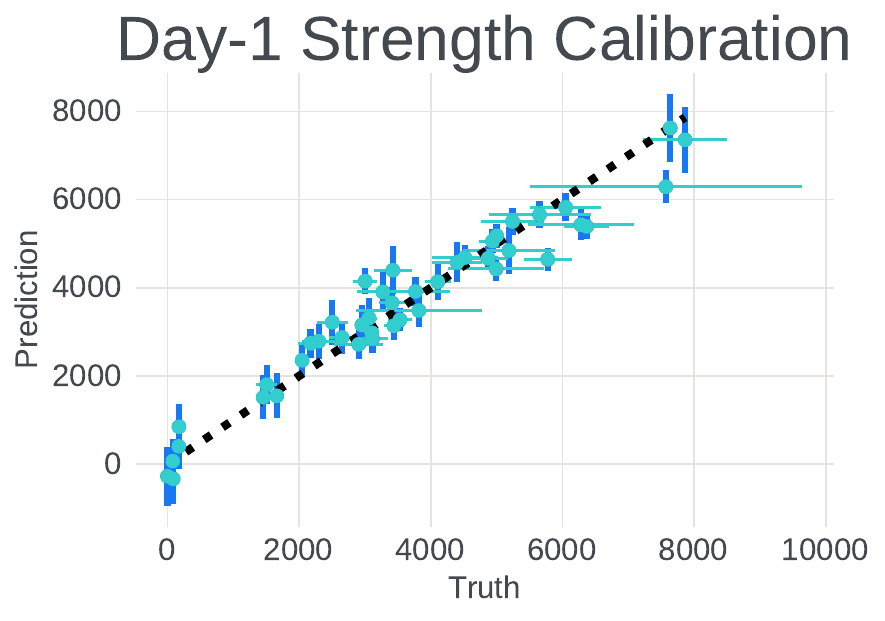}\vspace{-1.5ex}
    \caption{Cross-validation results for day-1 compressive strength predictions.}
    \label{fig:experiments:strength_calibration}
\end{figure}

\paragraph{Strength Model Cross-Validation}
Figure~\ref{fig:experiments:strength_calibration} shows cross validation results across all day-1 strength values in our data set, highlighting generally good predictive accuracies 
and well-calibrated uncertainties.

\paragraph{On monotonicity}
Another characteristic of concrete strength is its monotonic increase to a terminal value.
While techniques for including this constraint on the derivatives of the GP have been proposed, 
they introduce additional complexities that we have -- so far -- 
not found to be worth the potential increase in model performance.
The model's predictive mean is already monotone in our empirical observations, due to the other model components, like the log-time transform, the additive time-dependent component,
and a range of different measurement times in the training data.
Including this constraint could be lead to an increase in trust and uptake of the model by practitioners long term.

\end{document}